\def\year{2020}\relax
\documentclass[letterpaper]{article} % DO NOT CHANGE THIS
\usepackage{aaai20}  % DO NOT CHANGE THIS
\usepackage{times}  % DO NOT CHANGE THIS
\usepackage{helvet} % DO NOT CHANGE THIS
\usepackage{courier}  % DO NOT CHANGE THIS
\usepackage[hyphens]{url}  % DO NOT CHANGE THIS
\usepackage{graphicx} % DO NOT CHANGE THIS
\usepackage{graphicx}  % DO NOT CHANGE THIS
\usepackage{algorithmic}
\usepackage{algorithm}
\usepackage{macros}
\usepackage{subcaption}
\usepackage{amsmath}
\usepackage{amssymb}
\usepackage{bbm}

\title{Estimating Aggregate Properties In Relational Networks With Unobserved Data }
\author{Varun Embar\thanks{Equal contribution}\\ 
UC Santa Cruz\\ 
vembar@ucsc.edu 
\And Sriram Srinivasan\footnotemark[1]\\ 
UC Santa Cruz\\ 
ssriniv9@ucsc.edu 
\And Lise Getoor\\ 
UC Santa Cruz\\ 
getoor@soe.ucsc.edu}
\begin{document}

\maketitle

\begin{abstract}
Aggregate network properties such as cluster cohesion and the number of bridge nodes can be used to glean insights about a network's community structure, spread of influence and the resilience of the network to faults. %Several such aggregate properties have been proposed in literature. 
Efficiently computing network properties when the network is fully observed has received significant attention \cite{faust:book94,cook2006mining}, however the problem of computing aggregate network properties when there is missing data attributes has received little attention.
Computing these properties for networks with missing attributes involves performing inference over the network. 
%In case some of missing or unobserved network data, we need to infer the unobserved data before computing these properties. 
Statistical relational learning (SRL) and graph neural networks (GNNs) are two classes of machine learning approaches well suited for inferring missing attributes in a graph. 
In this paper, we study the effectiveness of these approaches in estimating aggregate properties on networks with missing attributes. 
We compare two SRL approaches and three GNNs.  
For these approaches we estimate these properties using point estimates such as MAP and mean.
For SRL-based approaches that can infer a joint distribution over the missing attributes, we also estimate these properties as an expectation over the distribution. 
To compute the expectation tractably for probabilistic soft logic, one of the SRL approaches that we study, we introduce a novel sampling framework. 
In the experimental evaluation, using three benchmark datasets, we show that SRL-based approaches tend to outperform GNN-based approaches both in computing aggregate properties and predictive accuracy. 
Specifically, we show that estimating the aggregate properties as an expectation over the joint distribution outperforms point estimates. 

\end{abstract}

\section{Introduction}
Large relational networks are ubiquitous, arising naturally across several domains such as social media (e.g., friendship and follower networks), computational biology (e.g., protein interaction networks) 
%the World Wide Web (e.g., hyperlink network) 
and IoT (e.g., sensor networks).
Structural properties, such as bridge nodes and graph clustering coefficients, are used to analyze the network for tasks such as influence maximization, community structure and spread of information.
%and resilience to attacks.  
%These techniques can be classified into compression-based \cite{navlakha2008graph,wu2014graph} sparsification-based \cite{shen2006visual,li2009egocentric}, influence-based \cite{shi2015vegas, qu2014interestingness} and aggregation-based techniques \cite{lefevre2010grass,dunne2013motif,adhikari2017condensing}.
%Among these, grouping or aggregation is the most popular summarization technique \cite{liu2018graph}.
While many such structural properties have been proposed 
%ss1: Varun: Below citation please verify if this is ok.
\cite{john:soc88,faust:book94,cook2006mining,rajaraman2011mining}, along with efficient algorithms to estimate them~\cite{shi:kde15,liu:cs18,wu:iciseee14,qu:ecml14,dunne:hfcs13}, the task of computing these properties in the presence of missing information, such as  node labels, has not received much attention.

In such networks, we need to combine the tasks of estimating these properties with inference of missing information. 
Here we examine two categories of network inference approaches: statistical relational learning (SRL) and graph neural networks (GNNs). 
SRL approaches \cite{getoor:book07} are one class of machine learning approaches which are well suited to making inferences in multi-relational networks. 
Most SRL approaches provide a way of specifying a declarative probabilistic model and efficient inference and learning algorithms \cite{richardson:ml06,bach:jmlr17,de:ilp08,friedman:ijcai99,neville:jmlr07}. In this paper, we investigate two such approaches, Markov logic networks (MLNs) \cite{richardson:ml06} and probabilistic soft logic (PSL) \cite{bach:jmlr17}. 
Both approaches specify a model using weighted logical rules, and use the model to define a joint distribution over the missing information. 
We can use the MAP or mean of this distribution to infer missing values, and estimate the properties, or we can compute the expectation of these properties over the joint distribution. 

GNNs are another class of machine learning approaches based on neural networks that can infer missing information in relational networks \cite{gilmer:icml17,hamilton:nips17,kipf:iclr16,velickovic:iclr18,qu:icml19}.
These approaches learn non-linear representations of nodes in the network and use these node representations to infer missing information.
Graph convolution networks (GCN) \cite{kipf:iclr16} and Graph attention networks (GAT) \cite{velickovic:iclr18} are two such popular approaches. 
Once the node representations are learned, these approaches independently infer the missing information for each node. 
Graph Markov neural networks (GMMN) \cite{qu:icml19} is a recently proposed neural network-based approach that models dependencies in the missing information along with node representations.
We can use these models to infer the missing values and estimate the properties. 
%However, none of these approaches infer a joint distribution over the unobserved data. 
We cannot compute the expectation for these approaches, as the final softmax layer infers a distribution for each node independently, and not the joint distribution over all nodes. 

In this work, we study the effectiveness of SRL and GNN based approaches in computing aggregate structural properties of networks with missing information. 
For the MAP and mean estimates of SRL approaches and for GNN based approaches, we use the point estimates to infer the missing information and then compute the aggregate property. 
For SRL approaches, we also compute the properties as an expectation over the joint distribution. 

Further, we introduce a novel sampling approach for computing expectation in PSL.
We propose an efficient Gibbs sampling based approach, \emph{ABGibbs}, to generate samples from the PSL joint distribution.
We use the generated samples to compute the expectation of aggregates tractably using Monte Carlo approximation.
\emph{ABGibbs} identifies highly correlated random variables (RVs), called \emph{association blocks}, and performs block sampling on these blocks.
This overcomes the poor performance of a naive Gibbs sampler due to high correlation between RVs. 

The contributions of our paper include:
1) We define several practical aggregate properties of a network; 
2) We propose a novel Metropolis-within-Gibbs sampling framework, \emph{ABGibbs}, to generate samples from the joint distribution of PSL; 
3) We analyze the effectiveness of three popular graph neural networks (GCN, GAT, GMNN) and two SRL methods (PSL and MLN) in computing the proposed aggregate properties; 
4) Through experiments on three benchmark datasets, we show that computing aggregate properties as an expectation outperforms point estimate, and the runtime experiments show that the proposed ABGibbs approach for PSL is up to 3 times faster than MLN sampling approach;

\commentout{
    \begin{itemize}
        \item {We define several practical aggregate properties of a network.}
        \item {We propose a novel Metropolis-within-Gibbs sampling framework, \emph{ABGibbs}, to generate samples from the joint distribution of PSL.}
        \item {We analyze the effectiveness of three popular graph neural networks (GCN, GAT, GMNN) and two SRL methods (PSL and MLN) in computing the proposed aggregate properties.}
        \item {Through experiments on three benchmark datasets, we show that computing aggregate properties as an expectation outperforms point estimate, and the runtime experiments show that the proposed ABGibbs approach for PSL is up to 3 times faster than MLN sampling approach.}
    \end{itemize}

}
\section{Background}
In this section, we first review the field of Statistical Relational Learning (SRL), including MLNs and PSL, followed by three graph neural network approaches(GNNs), GCN, GAT and GMNN.

\subsection{Statistical Relational Learning}
SRL methods combine probabilistic reasoning with knowledge representations that capture the structure in relational data. 
SRL frameworks define a joint probability distribution over the set of all possible networks using a declarative probabilistic model. 
An SRL model  $M$ is defined by a set of first order formula $F_{i}$, associated with weights $w_{i}$. Intuitively, the weight of a formula indicates how likely it is that the formula is true in the world. 
%Higher the weight, higher is the probability of formula being true. 

Given the domain for the variables in the formulas, SRL approaches generate a set of ground formulas, where the variables are replaced with the values in the domain. 
The atoms in the formula, where the variables are replaced with the values, are called ground atoms. 
These approaches then induce an undirected graphical model over the set of ground atoms, where each ground atom is modeled as a random variable (RV). The cliques in the graphical model correspond to the ground formulas.

Given an assignment to a set of ground atoms $X$, the probability distribution over the remaining unobserved ground atoms $Y$ is given by:
\begin{equation}
    P(Y = y| X = x; w) = \frac{1}{Z} exp \left( \sum_{i=1}^m w_{i}\phi_{i}(x,y) \right) 
\end{equation}
where $\phi_{i}(x, y)$ are the potential functions corresponding to the ground formulas and $Z$ is the normalization constant. 

\textbf{Markov Logic Networks (MLNs):}  MLNs \cite{richardson:ml06} are a notable SRL framework where the ground atoms are modeled as binary RVs. 
The potential functions are defined using boolean satisfiability, and take the value 1 if the ground formula is satisfied, and 0 otherwise.  

\textbf{Probabilistic Soft Logic:} PSL\cite{bach:jmlr17} is another recently introduced SRL framework. 
Unlike MLNs, the ground atoms in PSL are continuous and defined over the range $[0,1]$, and the weights $w_i$ are restricted to $\mathbf{R}^+$.
For the potential functions, PSL uses a continuous relaxation of boolean logic which results in  hinge functions instead of boolean satisfiability. 
These differences make MAP inference in PSL convex. 

\subsection{Graph neural networks}
Another line of research for inferring missing information in networks is based on the recent progress of graph neural networks. 
These techniques learn a non-linear representation for each node using neural networks and use these representations to infer missing attributes independently. 

\textbf{Graph Convolution Networks (GCNs):} GCNs \cite{kipf:iclr16} iteratively update the representation of each node by combining each node's representation with its neighbors' representation. 
The propagation rule to update the hidden representation of a node is given by:
\begin{equation}
    H^{(l+1)} = \sigma \left( \tilde{D}^{-0.5}\tilde{A}\tilde{D}^{-0.5}H^{(l)}W^{(l)} \right)  
\end{equation}
where $H^{(l)}$ denotes the representation at layer $l$, $\tilde{D}$ represents the degree matrix, $\tilde{A}$ represents the adjacency matrix with self-loop, and $W$ represents the weights. 
$\sigma$ denotes an activation function, such as the ReLU. 
The final representations are fed into a linear softmax layer classifier for label prediction. 

\textbf{Graph Attention Networks (GATs):} GATs \cite{velickovic:iclr18} are similar to GCNs, where node representations are updated iteratively by combining the representation of each node with its neighbors. 
However, instead of using a graph Laplacian, GATs use self-attention while combining representations. This allows the model to assign different weights to each of its neighbors' representations. 
The propagation rule for GAT is given by:
\begin{equation}
    h_i^{(l+1)} = \sigma \left( \sum_{j \in \mathcal{N}}\alpha_{ij} h_i^{(l)}W \right)  
\end{equation}
where $h_i^{(l)}$ represents the representation of node $i$ at layer $l$, $W$ represents the weight matrix and $\alpha$ represents the attention weights.

\textbf{Graph Markov Neural Networks (GMNNs):} GMNNs \cite{qu:icml19} build on graph neural networks such as GCNs or GATs. 
They add a second neural network to capture the latent dependencies in the inferred data.  
The pair of neural networks are trained using a variational EM algorithm. 
In the E-step, the object representations are learned by the first neural network.
In the M-step, the latent dependencies are learned by the other neural network. 
\section{Problem Definition}
Consider a network $G = (V,\mathcal{E})$, where $V$ is the set of nodes
and $\mathcal{E}$ is the set of edges. 
Each node $v_{i} \in V$ in the graph is associated with a set of attributes denoted by $\mathbf{a}_{v_{i}}$.
We assume that all nodes and edges are observed.
However, the set of node attributes may be incomplete, with some node attributes unobserved in the data. 
We denote the observed attributes of a node $v_{i}$ by $\mathbf{a}_{v_{i}}^{o}$, and the set of unobserved attributes by $\mathbf{a}_{v_{i}}^{u}$.

The goal is to estimate an aggregate property of the graph, $f(G, \mathbf{a}_{v_{i}})$, that involves the nodes, edges and the node attributes. 
These aggregate properties typically involve computing the sum, average, count, etc., on a set of nodes, edges and attributes that satisfy certain conditions.
We assume that the aggregate property is a real number, i.e., $f: (G, \mathbf{a}_{v_{i}}) \rightarrow \mathbbm{R}$. 
The function $f$ cannot be estimated directly due to the unobserved attributes $\mathbf{a}_{v_{i}}^{u}$. 
We need to infer these attributes before estimating the property $f$. 

One approach to estimate $f$, which we refer to as the \textit{Attribute Point Estimate approach}, is to impute the \textit{best} possible value for the unobserved attributes and then evaluate the property $f$. 
In this approach, we first learn a model by minimizing an objective function, such as the likelihood or loss, using the observed attributes $\mathbf{a}_{v_{i}}^{o}$ and impute values to the missing attributes using the learned model.
This approach can be denoted by:
\begin{equation}
    f(G, \mathbf{a}_{v_{i}}) = f(G, \hat{a}^u_{v_{i}}, \mathbf{a}^{o}_{v_{i}}) 
\end{equation}
where $\hat{a}^{u}_{v_{i}}$ denotes the imputed values for the unobserved attributes. 

Another approach, which we refer to as the \textit{Expected Aggregate} approach, is to model the unobserved attributes as RVs, define a joint probability distribution over them, and take the expectation of the property $f$ over the distribution. 
Since the range of $f$ is $\mathbbm{R}$, the expectation is well-defined. 
This approach can be denoted by:
\begin{equation}
    f(G, \mathbf{a}_{v_{i}}) = \mathbbm{E}_{p(\mathbf{a}_{v_{i}}^{u}|G, \mathbf{a}^o_{v_i})} [f(G, \mathbf{a}_{v_{i}}^{u}, \mathbf{a}^{o}_{v_i}) ]
\end{equation}
where $p(\mathbf{a}_{v_{i}}^{u} |G,\mathbf{a}^o_{v_i})$ denotes the probability distribution over the missing node attributes.
\section{Probabilistic aggregate estimation}
SRL techniques such as PSL and MLNs model missing node attributes as RVs and define a joint probability distribution over them. 
This makes the aggregate properties a function of RVs and we can compute the expected value of these functions over the distribution. 
However, computing the expectation analytically may not always be possible due the intractability of the integration in the expectation. 

One way to overcome this problem is through the use of Monte Carlo methods to approximate the expectation by drawing samples from the distribution. 
The expectation can be approximated as follows:
\begin{equation}
    f(G, \mathbf{a}_{v_{i}}) \approx \frac{1}{S} \sum_{j = 1}^{S} f(G, \mathbf{a}_{v_{i}}^o, \mathbf{a}_{v_{i}}^{u(j)})
\end{equation}
where $\mathbf{a}_{v_{i}}^{u(j)}$ are samples drawn from the distribution $p(\mathbf{a}_{v_{i}}^u | G, \mathbf{a}_{v_{i}}^o)$. 

Gibbs sampling \cite{gilks:book95} is a type of MCMC sampling approach that generates samples from the joint distribution by iteratively sampling from the conditional  distribution of each RV keeping the remaining RVs fixed. 
For MLNs, approaches such as MC-SAT have been proposed \cite{poon:aaai06},that combine MCMC and satisfiability, and are shown to greatly outperform Gibbs sampling.

However, 
%no approach exists for PSL to generate samples from the joint distribution. 
using Gibbs sampling for PSL has two main challenges. 
First, unlike MLNs, where the conditional distributions follow a binomial distribution and is easy to sample, it is non-trivial to generate samples from the conditional distributions of PSL. 
The conditional distribution for a RV $y_{i}$ conditioned on all other variables $X$, $Y_{-i}$ in PSL is given by:
\begin{equation}
\label{eq:conditional}
p(y_{i}|X,Y_{-i}) \propto exp\{ \sum_{r=1}^{N_{i}} w_{r} \phi_{r}(y_{i}, X, Y_{-i})\}
\end{equation}
where $N_{i}$ is the number of groundings that variable $y_{i}$ participates in. 
The above distribution neither corresponds to a standard named distribution nor has a form amenable to techniques such as inversion sampling. 
Second, Gibbs sampling has poor convergence rates when the RVs are highly correlated. 
Identifying such high probability regions is challenging.

Unlike previously existing hit-and-run based sampling approach \cite{broecheler:nips10}, we propose a simple and effective Gibbs sampling based approach to handle both these challenges. 
Our proposed sampling approach, \textbf{ABGibbs}, overcomes the first challenge of sampling from the conditional by incorporating a single step of a Metropolis algorithm within the Gibbs sampler (also called Metropolis-within-Gibbs \cite{gilks:book95}). 
For each RV $y_{i}$, we first sample a new value $y_{i}'$ from a symmetric proposal distribution $g(y_{i})$ and compute the acceptance ratio $\alpha$ given by:
\begin{equation}
\alpha = \frac{exp\{ \sum_{r=1}^{N_{i}} w_{r} \phi_{r}(y_{i}', X, Y_{-i})\}}{exp\{ \sum_{r=1}^{N_{i}} w_{r} \phi_{r}(y_{i}, X, Y_{-i})\}}
\end{equation}
We then accept the new value $y_{i}'$, as a sample from the conditional, with a probability proportional to the acceptance ratio $\alpha$. 

Highly weighted PSL rules with multiple unobserved terms result in highly correlated RVs. 
In general, highly weighted rules with more than two unobserved atoms can result in a large fraction of RVs becoming highly correlated, making it very hard to sample from the distribution. 
However, in the case of rules with up to two unobserved atoms, we propose a novel strategy to identify and cluster the correlated variables. We refer to the groups that we construct as
%SS: wanted to refer to workshop paper, but that might give away the author names. So thought will edit that paper and make sure only some sections and extra results from workshop can be submitted.
\textit{associated blocks}. 
%Moreover, the region of high probability for RVs in the cluster is given by a set of constraints on the sum and differences between pairs of RV.  
%ABGibbs identifies associated blocks based on the PSL program and the set of constraints. 
The primary idea of ABGibbs is to combine association blocks and Metropolis-within-Gibbs technique to generate samples efficiently from the joint distribution of a PSL model. 
%At a high-level, PSL rules that generate highly correlated pairwise terms are identifies and clusters of RVs are formed. 
%Then the values for the RVs in a block is jointly sampled in every iteration and are accepted or rejected as a block.
The algorithm for generating samples is shown in \algoref{algo:sampler}.
Our approach first identifies association blocks using the algorithm shown in \algoref{algo:findeqclass}.
We identify pairwise associations between ground atoms from ground rules with weights greater than a threshold $\lambda_{t}$.
We also keep track of the region where these potential functions are minimized.
This corresponds to a region of high probability.
We identify ground atom pairs where the region of high probability is below a threshold $\theta$.
Finally, we merge these pairs into blocks such that all associated pairs lie in the same block.

\begin{algorithm}[ht]
\caption{ABGibbs sampler for PSL}
\label{algo:sampler}
\begin{algorithmic}
\STATE \textbf{Input:} Set of $N$ ground rules $R$, \# of iterations T, burn-in period $b$, weight threshold $\lambda_{t}$, range threshold $\theta$.
\STATE \textbf{Output:} Set of samples $\mathcal{S}$
\STATE \# Identify the associated blocks using the ABlock algorithm
\STATE $\mathcal{C} \gets$ABlocks($R$, $\lambda_{t}$, $\theta$)
\STATE \# Initialize $Y^{(0)}$ to MAP state
\STATE $Y^{(0)} \gets {arg max}_{Y} p(Y|X)$
\FOR{$t$ from  $1$ to $T$}
    \STATE \# Sample values for each block $c \in \mathcal{C}$
	\FOR{$c^{(t)} \in \mathcal{C}^{(t)}$}
		\STATE $c' \sim BlockSample(c, R+, R-)$
		\STATE $\alpha = \frac{exp\{ \sum_{r=1}^{N_{i}} w_{r} \phi_{r}(c', X, Y_{\setminus c}^{(t+1)})\}}{exp\{ \sum_{r=1}^{N_{i}} w_{r} \phi_{r}(Y_{c}^{(t)}, X, Y_{\setminus c}^{(t+1)})\}}$
		\STATE $u \sim \mathbf{U}(0,1)$
		\IF {$u < \alpha$}
			\STATE $Y_{c}^{(t+1)} = c'$
		\ELSE
			\STATE $Y_{c}^{(t+1)} = Y_{c}^{(t)}$
		\ENDIF
	\ENDFOR
	\STATE \# Consider samples after burn-in period $b$
	\IF {$t > b$}
		\STATE $\mathcal{S} = \mathcal{S} \cup Y^{(t)}$
	\ENDIF
\ENDFOR
\STATE Return $\mathcal{S}$
\end{algorithmic}
\end{algorithm}

\begin{algorithm}[ht]
\caption{ABlock: Identifying blocks of associated RVs}
\label{algo:findeqclass}
\begin{algorithmic}
\STATE \textbf{Input:} Set of $N$ ground rules $G$, weight threshold $\lambda_{t}$, range threshold $\theta$
\STATE \textbf{Output:} Blocks of associated RVs(RVs) $\mathcal{C}$
\STATE \textbf{Initialize:} Hashmaps $R^{+}$ and $R^{-}$ that hold additive and subtraction bounds
\FOR{$r \in 1\ to\ N$}
	\STATE \# For rules with high weights
	\IF{$\lambda_{r} > \lambda_{t}$}
		\STATE \# Update the bounds
		\IF{$r$ is of the form $a - b \leq c$}
			\STATE ${R}^{-}(a,b).max$ = $\min\{R^-(a,b).max, c\}$
		\ELSIF{$r$ is of the form $a - b \geq c$}
			\STATE ${R}^{-}(a,b).min = \max\{R^-(a,b).min, c\}$
		\ELSIF{$r$ is of the form $a + b \leq c$}
			\STATE ${R}^{+}(a,b).max = \min\{R^+(a,b).max, c\}$
		\ELSIF{$r$ is of the form $a + b \geq c$}
			\STATE ${R}^{+}(a,b).min = \max\{R^+(a,b).min, c\}$
		\ENDIF
	\ENDIF
\ENDFOR
\STATE \# Identify clusters from pairwise associations
\FOR{$(a,b) \in R^{+} \bigcup R^{-}$}
	\IF{${R}^{+}(a,b).max - {R}^{+}(a,b).min \leq \theta$ or ${R}^{-}(a,b).max - {R}^{-}(a,b).min \leq \theta$}
		\STATE{Merge blocks containing a,b and update $\mathcal{C}$}
	\ENDIF
\ENDFOR 
\STATE Add remaining RVs as singleton clusters to $\mathcal{C}$
\STATE Return set of blocks $\mathcal{C}$
\end{algorithmic}
\end{algorithm}

Having identified the association blocks, we generate samples from the distribution using a blocked Metropolis-in-Gibbs sampler. 
We initialize the RVs to the MAP state.
We then iteratively sample new values for each association block $B_{k}$ after a burn-in period $b$. 
The proposed sampling approach for each block is given in \algoref{algo:sample}.
We first randomly choose a variable $y_{i} \in B_{k}$ and sample a value from $\mathbf{U}(0,1)$.
We then update the region of high probability for all RVs in $B_{k}$ based on the sampled value for $y_{i}$.
We randomly choose a variable $y_{i+1} \in B_{k}$ such that $y_{i+1} \neq y_{1}, \dots, y_{i}$, and sample a value from the region of high probability with probability $\beta$ and sample a value from $Unif(0,1)$ with probability $1-\beta$.
We again update the region of high probability for all variables in $B_{k}$ with $y_{i+1}$.
This process is performed iteratively for all the variables in the block. 
Finally, once we have sampled a value for all the variables in a block, we accept or reject  the samples for all $y_{i} \in B_{k}$ with probability $\alpha$. 
A single sample from the joint distribution is complete when every block has been sampled once.

\begin{algorithm}[ht]
\caption{BlockSample:  Sampling variables in a block}
\label{algo:sample}
\begin{algorithmic}
\STATE \textbf{Input:} A block of RVs $\mathbf{c}$, ${R}^{+}, R^{-}$ 
\STATE \textbf{Output:} Sample $\mathbf{s}$ for variables in $\mathbf{c}$
%\STATE \#$\mathbf{s}$ is the set of sampled variables
\STATE $\mathbf{s}$ = $\emptyset$
\STATE Pick a variable $y_{i}$ from $\mathbf{c}$ at random
\STATE $y_{i} \sim \mathbf{U}(0,1)$ 
\STATE $\mathbf{s}.add(y_i)$
\WHILE{$y_{j} \in \mathbf{c} \setminus \mathbf{s}$ and associated to some variable in $\mathbf{s}$}
	\STATE Update range $[u,v]$ for $y_{j}$ based on $\mathbf{s}$, $R^{+}$, and $R^{-}$
	\STATE $b \sim [0,1]$
	\IF{ $b \leq \beta$}
		\STATE $y_{j} \sim \mathbf{U}(u,v)$
	\ELSE
		\STATE $y_{j} \sim \mathbf{U}(0,1)$
	\ENDIF
	\STATE $\mathbf{s}.add(y_{j})$
\ENDWHILE
\STATE Return $\mathbf{s}$
\end{algorithmic}
\end{algorithm}
\section{Aggregate properties} \label{sec:queries}
Next we discuss the several aggregate queries studied in this paper that are useful in analyzing community structure and spread of influence in social networks.
%We introduce useful aggregate queries over attributed networks. 
%Although we motivate these queries through a labeled citation network, these queries can be useful across other domains such as biological and social networks.
We illustrate using  a citation network given by $G = (V, \mathcal{E}, a_v)$, where $V = \{v_{1}, \dots, v_{n}\}$ correspond to documents, and 
$\mathcal{E} = \{e_{uw} | u,w \in V\}$, corresponds to a citation link from documents $u$ to $w$. 
For each node $v$, the set of node attributes is given by $a_v$. The node attribute corresponding to the document category is denoted by $c_{i} \in \{0,\dots, \kappa\}$ where $\kappa$ is the number of categories. 
On the above network, we define five different queries Q1 to Q5. Q1 and Q2 are inspired from cluster analysis \cite{tan:book06} and Q3, Q4, and Q5 are related to bridge nodes in social networks 
%ss1: Varun: I am not sure about the citation I added below, can you verify this please.
\cite{musia:cci09:}.

\textbf{[Q1]: Category Cohesion: } 
This property is defined as the count of document pairs $(v_{i}, v_{j})$ that have a citation link and belong to the same category $c_{i} = c_{j}$, i.e., 
\begin{align*}
Q1 = \sum_{e_{ij} \in \mathcal{E}}\mathbbm{1}(c_{i}= c_{j})
\end{align*}  
where $\mathbbm{1}$ is an indicator function with value one when the condition is satisfied.  
%ss1: Varun, Lise pointed out that there is a degenrate case in the statement below, we can address that by dividing everything by number of nodes. Should we do that?
%Networks consisting of categories that form well-separated clusters have high score for this property.
Network where categories are isolated have a high value. 

\textbf{[Q2]: Category Separation: } 
This property is defined as the count of document pairs $(v_{i}, v_{j})$ that have a citation link between them and belong to the different category $ c_{i} \neq c_{j}$, i.e.,
\begin{align*}
   Q2 = |\mathcal{E}| - Q1%\sum_{e_{ij} \in \mathcal{E}}\mathbbm{1}(c_{i} \neq c_{j})
\end{align*}
Networks that have related categories have a large value for this property.

\textbf{[Q3]: Diversity of Influence: } 
This property is defined as the number of documents $v_{i}$ in the network that are connected to documents belonging to more than half the categories, i.e., 
\begin{align*}
Q3 =& \sum_{i = 1}^{n}\mathbbm{1}\Bigg( 
&\Big|\{c_{j} \mid \forall_{j=1}^{n} \big(e_{ij} \in \mathcal{E} \land c_{i} \neq c_{j}\big)\}\Big| \geq \frac{\kappa}{2}\Bigg)
\end{align*}
where $|\{\dots\}|$ indicates number of elements in the set. 
Networks containing documents that have influenced many categories have a large value for this property. 
 
\textbf{[Q4]: Exterior Documents: } 
This property is defined by the number of documents $v_{i}$ that have more than half the neighbors belonging to categories other  than the documents category $c_{i}$, i.e.,
\begin{align*}
    Q4 = \sum_{i=1}^{n} \mathbbm{1}\Bigg(\Big(\sum_{j=1}^{n}e_{ij} \mathbbm{1}\big(c_{i} \neq c_{j} \big)\Big) > \frac{\sum_{j=1}^{n}e_{ij}}{2}\Bigg)
\end{align*}
Unlike Q3, Q4 measures the number of documents that have more reach in a different category than their own. 
Networks where the categories are not well separated typically have a large value for this property.

\textbf{[Q5]: Interior Documents: }
This property is defined as the number of documents $v_{i}$ that have more that half of its neighbors belonging to the same category as the document $c_{i}$, i.e.,
\begin{align*}
    Q5 = \sum_{i=1}^{n} \mathbbm{1}\Bigg(\Big(\sum_{j=1}^{n}e_{ij} \mathbbm{1}\big(c_{i} = c_{j} \big)\Big) > \frac{\sum_{j=1}^{n}e_{ij}}{2}\Bigg)
\end{align*}
Networks with large category clusters typically have a large value for this property.

\section{Empirical Evaluation}
In this section we analyze the performance of SRL and GNN-based approaches on the queries proposed in the previous section. 
We also compare the predictive and runtime performance of these approaches. 

\begin{table}[t]
\centering
\resizebox{0.4\textwidth}{!}{
\begin{tabular}{|c|c|c|c|c|c|}
\hline
Dataset & \#Classes & \#Nodes & \#Edges & \#Feats & \#Obs Labels \\ \hline
Cora & 7 & 2708 & 5429 & 1433  & 640 \\ \hline
Pubmed & 3 & 19717 & 44338 & 500  & 560\\ \hline
Citeseer & 6 & 3327 & 4732 & 3703 & 620\\ \hline
\end{tabular}
}
\caption{Statistics for the three datasets: Cora, Pubmed and Citeseer. }
\label{tab:dataset}
\end{table}

\begin{table*}[t]
\centering
\resizebox{\textwidth}{!}{
\begin{subtable}[t]{0.6\textwidth}
\centering
\caption{Cora}
\label{tab:cora}
\begin{tabular}{|c|c|c|c|c|c|c|c|}
\hline
Methods     & Q1 - $\delta$ & Q2 - $\delta$ & Q3 - $\delta$ & Q4 - $\delta$ & Q5 - $\delta$ & $\hat{\delta}$ \\ \hline
PSL-MAP     & 0.047      & 0.205      & 0.165      & 0.1        & 0.062       & 0.115      \\ \hline
MLN-MAP     & 0.032      & \textbf{0.046}      & 0.412       & 0.436      & 0.242       & 0.234      \\ \hline
PSL-MEAN    & 0.021      & 0.090      & 0.027      & 0.054      & 0.041       & 0.047      \\ \hline
MLN-MEAN    & 0.038      & 0.163      & \textbf{0.009}      & 0.174      & 0.068       & 0.090      \\ \hline
GCN         & 0.048      & 0.207      & 0.137      & 0.671      & 0.34        & 0.28       \\ \hline
GAT         & 0.073      & 0.313      & 0.376      & 0.697      & 0.355       & 0.362      \\ \hline
GMNN        & 0.071      & 0.306      & 0.174      & 0.711      & 0.352       & 0.322      \\ \hline \hline
PSL-SAMPLES & \textbf{0.014}      & 0.061      & 0.050      & \textbf{0.053}       & \textbf{0.031}       & \textbf{0.041}      \\ \hline
MLN-SAMPLES & 0.045      & 0.161       & 0.042      & 0.173      & 0.068       & 0.097      \\ \hline

\end{tabular}
\end{subtable}\ \ %
\begin{subtable}[t]{0.6\textwidth}
\centering
\caption{Pubmed}
\label{tab:pubmed}
\begin{tabular}{|c|c|c|c|c|c|c|}
\hline
Methods     & Q1 - $\delta$ & Q2 - $\delta$ & Q3 - $\delta$ & Q4 - $\delta$ & Q5 - $\delta$ & $\hat{\delta}$ \\ \hline
PSL-MAP     & 0.13       & 0.528      & 0.396      & 0.714      & 0.121       & 0.377      \\ \hline
MLN-MAP     & 0.109      & 0.491      & 0.281      & 0.570      & 0.102       & 0.310      \\ \hline
PSL-MEAN    & 0.117      & 0.474      & 0.348      & 0.685      & 0.115       & 0.347      \\ \hline
MLN-MEAN    & 0.064      & 0.261      & \textbf{0.113}      & \textbf{0.362}      & \textbf{0.053}       & 0.170      \\ \hline
GCN         & 0.089      & 0.361      & 0.169      & 0.626      & 0.102       & 0.269      \\ \hline
GAT         & 0.129      & 0.526      & 0.293      & 0.709      & 0.119       & 0.355      \\ \hline
GMNN        & 0.156      & 0.513      & 0.299      & 0.679      & 0.119       & 0.353      \\ \hline \hline
PSL-SAMPLES & 0.108      & 0.441      & 0.312      & 0.618      & 0.105       & 0.316      \\ \hline
MLN-SAMPLES & \textbf{0.060}      & \textbf{0.210}       & 0.119      & 0.391      & 0.061       & \textbf{0.168}      \\ \hline

\end{tabular}
\end{subtable}
}
\resizebox{0.8\textwidth}{!}{
\begin{subtable}[t]{\textwidth}
\centering
\caption{Citeseer}
\label{tab:citeseer}
\begin{tabular}{|c|c|c|c|c|c|c|}
\hline
Methods      & Q1 - $\delta$ & Q2 - $\delta$ & Q3 - $\delta$ & Q4 - $\delta$ & Q5 - $\delta$ & $\hat{\delta}$ \\ \hline
PSL-MAP        & 0.175      & 0.527      & 0.673      & 0.57       & 0.272       & 0.443      \\ \hline
MLN-MAP        & 0.207      & 0.648      & 0.594      & 0.794      & 0.392       & 0.527      \\ \hline
PSL-MEAN         & \textbf{0.134}      & \textbf{0.403}      & 0.544      & 0.551      & 0.253       & 0.377      \\ \hline
MLN-MEAN       & 0.137      & 0.731      & 0.792       & 0.691      & 0.315       & 0.554      \\ \hline
GCN          & 0.211      & 0.637      & 0.712      & 0.813      & 0.396       & 0.553      \\ \hline
GAT           & 0.248      & 0.747      & 0.9        & 0.887      & 0.416       & 0.639      \\ \hline
GMNN          & 0.257      & 0.774      & 0.881      & 0.906      & 0.447       & 0.653      \\ \hline \hline
PSL-SAMPLES   & 0.137      & 0.413      & \textbf{0.539}      & \textbf{0.499}       & \textbf{0.236}       & \textbf{0.364}      \\ \hline
MLN-SAMPLES  & 0.244      & 0.736      & 0.793       & 0.691      & 0.315        & 0.555      \\ \hline

\end{tabular}
\end{subtable}
}
\caption{Relative errors ($\delta$) for different queries on the three datasets. PSL-SAMPLES and MLN-SAMPLES have lower error. These approaches compute the expectation of the properties over the distribution.}
\label{tab:querypred}
\end{table*}

\subsection{Experimental Setup and Datasets}
We consider three benchmark citation datasets for relational learning: Cora, Pubmed and Citeseer \cite{sen:ai08}. 
The statistics for these datasets are given in \tabref{tab:dataset}. 
Each node in the network corresponds to a document, and has attributes describing the words occurring in a document represented as a bag-of-words, and an attribute that represents the category of the document. 
Links between documents represent citations. 
We assume all the words and citations are observed, while the categories are only partially observed.
We follow the same splits as \cite{yang:icml16} and the number of observed node labels is given in \tabref{tab:dataset}. 

%For transductive inference approaches (MLN and PSL), we use the test dataset as unlabeled. 

\noindent{\textbf{SRL approaches}: 
For both MLNs and PSL, we extend the model defined in \cite{bach:jmlr17} to incorporate the bag-of-words features.
We first train a logistic regression (LR) model with L2 regularization (with 0.01 as weight for regularizer) to predict the node labels using the bag-of-words features. 
For each node, we consider the category with the highest probability as the LR prediction. 
We use all the observed node labels to train the LR model. 

The model contains a general label propogation rule of the form:
$w: HasCat(A, C) \land Link(A, B) => HasCat(B, C)$.
The model also contains a category specific label propagation rule of the form: 
$w: HasCat(A,  $`c'$ ) \land Link(A, B) => HasCat(B, $`c'$)$
for each category `c'.
The model incorporates LR predictions using the rule of the form:
$w: LR(a, $`c'$) \rightarrow HasCat(a, $`c' $)$.
For MLN, we include a functional constraint that prevents a node from having multiple categories set to true.
For PSL, we include a highly weighted rule that states that the truth values across all categories must sum to 1 for a node.
We perform weight learning using MC-SAT for MLN and maximum likelihood estimation for PSL.}

The different SRL based approaches that we consider are:
\noindent{\textbf{MLN-MAP}: 
This is the mode of the distribution defined by the MLN model over the unobserved node labels. We use the MaxWalkSAT algorithm implemented in the Tuffy framework \cite{niu2011tuffy}.}

\noindent{\textbf{MLN-MEAN}: This is mean of the distribution defined by the MLN model over the unobserved node labels.
We generate 1100 samples using the MC-SAT algorithm, discard the first 100 samples as burn-in samples and use the 1000 samples to approximate the mean.}

\noindent{\textbf{MLN-SAMPLES}: Here we estimate the properties as an expectation over the distribution defined by the MLN model. We generate samples similar to MLN-MEAN, but randomly choose 100 samples from the 1000 (to ensure minimal correlation) and use Monte Carlo approximation to compute aggregate property expectation.}

\noindent{\textbf{PSL-MAP}: This is the mode of the distribution defined by the PSL model over the unobserved node labels. We use ADMM algorithm implemented in the PSL framework \cite{bach:jmlr17} to obtain labels.} 

\noindent{\textbf{PSL-MEAN}: This is mean of the distribution defined by the PSL model over the unobserved node labels.
We generate 1100 samples using the proposed ABGibbs algorithm, discard the first 100 samples as burn-in samples and use the 1000 samples to approximate the mean.}

\noindent{\textbf{PSL-SAMPLES}: Here we estimate the properties as an expectation over the distribution defined by the PSL model. Like MLN-SAMPLE we generate 100 samples and use our proposed  Monte Carlo approximation to compute aggregate property expectation} \\

\noindent{\textbf{GNN based approaches:} The approaches use the node representations to infer node labels. 
These models use 20 observed node labels from each category to train the model and use the remaining 500 node labels for performing early-stopping. For all three approaches we use the code and hyperparameters provided by the authors of the respective paper. The different GNN based approaches that we consider are: }

\noindent{\textbf{GCN}: This approach returns a point estimate computed by the graph convolutional network \cite{kipf:iclr16}.}

\noindent{\textbf{GAT}: This approach returns a point estimate computed by the graph attention network \cite{velickovic:iclr18}.}

\noindent{\textbf{GMNN}: This approach also returns a point estimate computed by the Markov neural network introduced recently \cite{qu:icml19}.}

We evaluate the performance on all the queries (Q1 to Q5) using the relative error ($\delta$) as a metric.
The relative error $\delta$ is computed using: $\delta = \frac{|P-T|}{T}$ where $T$ is the true value of the query and $P$ is the estimated value. 
We evaluate the overall performance of a method by computing the mean over all the $\delta$s denoted by $\hat{\delta}$. 
We also report the predictive performance of these methods, by computing the categorical accuracy (Acc) on all the unobserved nodes. 
Further, for runtime comparison, we measure the total time taken for each of these approaches. 
    
    \subsection{Predictive Performance}
    
    \begin{table}[h]
    \centering
    \resizebox{0.35\textwidth}{!}{
    \begin{tabular}{|c|c|c|c|}
    \hline
    Methods     & \begin{tabular}[c]{@{}c@{}}Cora\\ Acc (\%)\end{tabular} & \begin{tabular}[c]{@{}c@{}}Pubmed\\ Acc (\%)\end{tabular} & \begin{tabular}[c]{@{}c@{}}Citeseer\\ Acc (\%)\end{tabular} \\ \hline
    PSL-MAP     & \textbf{85.34} & \textbf{83.6} & \textbf{72.25} \\ \hline
    MLN-MAP     & 77.9           & 76.75           & 71.7         \\ \hline
    PSL-MEAN    & 84.13          & 83.16          & 71.7         \\ \hline
    MLN-MEAN    & 82.35          & 75.14          & 71.25         \\ \hline
    GCN         & 81.96          & 77.73          & 68.78         \\ \hline
    GAT         & 81.43          & 76.87          & 70.41         \\ \hline
    GMNN        & 83.26          & 81.07          & 70.15         \\ \hline \hline
    PSL-SAMPLES & 83.01          & 81.88          & 71.29         \\ \hline
    MLN-SAMPLES & 82.25          & 73.48          & 71.11         \\ \hline
    \end{tabular}
    }
    \caption{Accuracy for the three datasets computed over the unobserved node labels. PSL-MAP has the best accuracy.}
    \label{tab:accuracy}
    \end{table}
    
    The accuracy for the node labeling task computed over the unobserved node labels is given in \tabref{tab:accuracy}. 
    We observe that PSL-MAP obtains the highest accuracy on all three dataset. 
    Further, we observe that for PSL, PSL-MAP, which is the mode of the distribution has higher accuracy than PSL-MEAN, which is the distribution mean. 
    However, for MLN, the mean of the distribution is better than the mode for Cora and Citeseer. 
    For both PSL and MLN, we observe that the mean is better than the average accuracy of the samples. 
    This is because samples with lower accuracy are also sampled (with low probability), which brings down the average. 
    %This might be due to the binary nature of MLN which forces nodes to take either zero or one. 
    Among the neural network based methods, we observe that GMNN performs the best. 
    This is due the ability of GMNN to model the dependencies in node labels.
    
    %A key thing to note when measuring the predictive power using a metric like accuracy is that, accuracy does not consider the strong relationships that the nodes share.
    %Accuracy considers the performance on each node independently and hence every mistake in node is peanalized equally. 
    %However, this is not true when it comes different query under consideration which we examine in the next section. 
    \subsection{Query Performance}
    We next analyze the performance of these approaches on the task of estimating aggregate properties on these  datasets. 
    The relative error for Cora is shown in \tabref{tab:cora}, Pubmed in \tabref{tab:pubmed}, and for Citeseer in \tabref{tab:citeseer}. 
    We observe that approaches that take the expectation over the aggregate properties perform better than point estimate based approaches.
    For Cora and Citeseer, PSL-SAMPLES outperforms all other approaches overall and on most queries independently, and 
    similarly MLN-SAMPLES outperforms all other approaches for Pubmed. 
    Overall PSL-MEAN and MLN-MEAN tend to be close to PSL-SAMPLES and MLN-SAMPLES as the means are computed using these samples.
    
    We next observe that for both PSL and MLNs, the SAMPLES usually have a much lower error when compared to the MAP estimates. 
    This is in contrast to the accuracy, where the SAMPLES do not perform as well as MAP estimates.
    The MAP, which corresponds to the mode of the distribution, assigns node labels such that  joint probability distribution is maximized.  
    However, for documents that lie in between multiple category clusters, the correct category assignment might have slightly lower, but still significant, probability mass. 
    Unlike accuracy, where all node labels are equally important, the nodes that lie in the border of the category clusters have a higher weight in the queries. 
    Since samples from the distribution contain the node labels proportional to the probability mass, PSL-SAMPLES and MLN-SAMPLES tend to perform better than their MAP counterparts. 
    
    Among the neural network based approaches we observe that GCN performs better than GAT and GMNN. 
    This is again in contrast to the accuracy, where GCN performs poorly, and GMNN has the higher Acc.

    Among the queries, we observe that Q1 and Q5 have lower error compared to the other queries for all the methods. 
    Both Q1 and Q5, estimate node pairs that are adjacent and have the same category. 
    These are easier to estimate as these nodes typically lie at the center of a category clusters. 
    As a result, the node attributes for these nodes and their adjacent nodes are similar. 
    Since all the approaches propagate the similarity between the node neighbors, the models have a lower error on these queries. 
    We computed the accuracy for nodes that participate in these queries and found that most of the methods had an accuracy of over 90\%.
    
    Queries Q2, Q3, and Q4 estimate nodes that have neighbors with different categories. 
    These are nodes that lie in the boundary of category clusters and whose categories are harder to infer. 
    We observed a reduction in the accuracy to about 60\% for nodes that participate in these queries. 
    Approaches such as GMNN have very large relative error for these queries, resulting in overall poor performance.
    
    %Finally, we observe that Bayesian-based approaches due to its robustness 
    %performs better on a variety queries that try to measure the prediction 
    %on the network as a whole. This performance however comes at the cost 
    %overall runtime. In the next section we discuss different runtimes 
    %required for different approaches.
    
    \subsection{Runtime Comparisons}
    \begin{table}[ht]
    \centering
    \resizebox{0.35\textwidth}{!}{
    \begin{tabular}{|c|c|c|c|}
    \hline
    Methods & \begin{tabular}[c]{@{}c@{}}Cora\\ Time (sec)\end{tabular} & \begin{tabular}[c]{@{}c@{}}Pubmed\\ Time (sec)\end{tabular} & \begin{tabular}[c]{@{}c@{}}Citeseer\\ Time (sec)\end{tabular} \\ \hline
    PSL-MAP &  14 & 124 & 37\\ \hline
    PSL-MEAN &  105 & 638  & 124\\ \hline
    MLN-MEAN &  270 &  1947 & 166\\ \hline
    MLN-MAP &  65 & 368  & 36\\ \hline
    GCN &  24 & 59 & 29 \\ \hline
    GAT &  142 & 138 & 122 \\ \hline
    GMNN &  30 & 17 & 8\\ \hline \hline
    PSL-SAMPLES &  105 & 638 & 124\\ \hline
    MLN-SAMPLES &  270 & 1947 & 166 \\ \hline
    \end{tabular}%
    }
    \caption{Table showing runtimes for each of the approaches on the three datasets.}
    \label{tab:timing}
    \end{table}
    In \tabref{tab:timing}, we show the runtime for each 
    of the approaches. 
    We observe that approaches that compute a point estimate are significantly faster  compared to sample-based approaches. 
    This is expected as point estimates are computed using efficient optimization approaches. 
    Not surprisingly, sampling-based approaches are not as efficient, however their runtimes are still reasonable.
    %This holds for both graph neural networks and MAP estimates of probabilistic approaches.
    The runtimes for MEAN and SAMPLES are the same, as we need to generate samples from the distribution for each of these approaches.
    Further, among PSL and MLN samplers, we observe that PSL  is faster by a factor of two for Cora and three for Pubmed.

\section{Conclusion and Future Work}
In this paper, we studied the task of estimating aggregate properties for networks with unobserved data. 
By comparing three graph neural networks and two probabilistic approaches on benchmark datasets, we show computing the expectation of aggregate properties over the distribution of unobserved random variables reduces the relative error. 
We have also proposed a blocked Gibbs sampling framework for PSL, that identifies pairwise correlated RVs and block samples them. 
We also showed the overall effectiveness of our approach through experiments. 

An interesting future direction is to combine graph neural network approaches with SRL models that can learn node representations and also infer a joint distribution over the unobserved data. 
Extending this analysis for networks with missing edges and nodes is another line of future work.

\section{Acknowledgements}
This work was partially supported by the National Science
Foundation grants CCF-1740850 and IIS-1703331, AFRL
and the Defense Advanced Research Projects Agency. Golnoosh Farnadi is supported by postdoctoral scholarships
from IVADO through the Canada First Research Excellence
Fund (CFREF) grant.
%Extending the ABGibbs approach to PSL program containing high weighted rules with more than two random variables is another line of future work.

\bibliography{reference}

\begin{thebibliography}{}

\bibitem[\protect\citeauthoryear{Bach \bgroup et al\mbox.\egroup
  }{2017}]{bach:jmlr17}
Bach, S.~H.; Broecheler, M.; Huang, B.; and Getoor, L.
\newblock 2017.
\newblock Hinge-loss markov random fields and probabilistic soft logic.
\newblock {\em Journal of Machine Learning Research} 18:1--67.

\bibitem[\protect\citeauthoryear{Broecheler and
  Getoor}{2010}]{broecheler:nips10}
Broecheler, M., and Getoor, L.
\newblock 2010.
\newblock Computing marginal distributions over continuous markov networks for
  statistical relational learning.
\newblock In {\em Advances in Neural Information Processing Systems}.

\bibitem[\protect\citeauthoryear{Cook and Holder}{2006}]{cook2006mining}
Cook, D.~J., and Holder, L.~B.
\newblock 2006.
\newblock {\em Mining graph data}.
\newblock John Wiley \&\#38; Sons, Inc.

\bibitem[\protect\citeauthoryear{De~Raedt and Kersting}{2008}]{de:ilp08}
De~Raedt, L., and Kersting, K.
\newblock 2008.
\newblock Probabilistic inductive logic programming.
\newblock In {\em Probabilistic Inductive Logic Programming}.

\bibitem[\protect\citeauthoryear{Dunne and Shneiderman}{2013}]{dunne:hfcs13}
Dunne, C., and Shneiderman, B.
\newblock 2013.
\newblock Motif simplification: improving network visualization readability
  with fan, connector, and clique glyphs.
\newblock In {\em Human Factors in Computing Systems}.

\bibitem[\protect\citeauthoryear{Friedman \bgroup et al\mbox.\egroup
  }{1999}]{friedman:ijcai99}
Friedman, N.; Getoor, L.; Koller, D.; and Pfeffer, A.
\newblock 1999.
\newblock Learning probabilistic relational models.
\newblock In {\em International Joint Conference on Artificial Intelligence}.

\bibitem[\protect\citeauthoryear{Getoor and Taskar}{2007}]{getoor:book07}
Getoor, L., and Taskar, B.
\newblock 2007.
\newblock {\em Introduction to Statistical Relational Learning}.
\newblock The MIT Press.

\bibitem[\protect\citeauthoryear{Gilks, Richardson, and
  Spiegelhalter}{1995}]{gilks:book95}
Gilks, W.~R.; Richardson, S.; and Spiegelhalter, D.
\newblock 1995.
\newblock {\em Markov chain Monte Carlo in practice}.
\newblock Chapman and Hall/CRC.

\bibitem[\protect\citeauthoryear{Gilmer \bgroup et al\mbox.\egroup
  }{2017}]{gilmer:icml17}
Gilmer, J.; Schoenholz, S.~S.; Riley, P.~F.; Vinyals, O.; and Dahl, G.~E.
\newblock 2017.
\newblock Neural message passing for quantum chemistry.
\newblock In {\em International Conference on Machine Learning}.

\bibitem[\protect\citeauthoryear{Hamilton, Ying, and
  Leskovec}{2017}]{hamilton:nips17}
Hamilton, W.; Ying, Z.; and Leskovec, J.
\newblock 2017.
\newblock Inductive representation learning on large graphs.
\newblock In {\em Advances in Neural Information Processing Systems}.

\bibitem[\protect\citeauthoryear{Kipf and Welling}{2017}]{kipf:iclr16}
Kipf, T.~N., and Welling, M.
\newblock 2017.
\newblock Semi-supervised classification with graph convolutional networks.
\newblock In {\em International Conference on Learning Representations}.

\bibitem[\protect\citeauthoryear{Liu \bgroup et al\mbox.\egroup
  }{2018}]{liu:cs18}
Liu, Y.; Safavi, T.; Dighe, A.; and Koutra, D.
\newblock 2018.
\newblock Graph summarization methods and applications: A survey.
\newblock {\em Computing Surveys} 51:62.

\bibitem[\protect\citeauthoryear{Musia{\l} and
  Juszczyszyn}{2009}]{musia:cci09:}
Musia{\l}, K., and Juszczyszyn, K.
\newblock 2009.
\newblock Properties of bridge nodes in social networks.
\newblock In {\em Computational Collective Intelligence. Semantic Web, Social
  Networks and Multiagent Systems}.

\bibitem[\protect\citeauthoryear{Neville and Jensen}{2007}]{neville:jmlr07}
Neville, J., and Jensen, D.
\newblock 2007.
\newblock Relational dependency networks.
\newblock {\em Journal of Machine Learning Research} 8:653--692.

\bibitem[\protect\citeauthoryear{Niu \bgroup et al\mbox.\egroup
  }{2011}]{niu2011tuffy}
Niu, F.; R{\'e}, C.; Doan, A.; and Shavlik, J.
\newblock 2011.
\newblock Tuffy: Scaling up statistical inference in markov logic networks
  using an rdbms.
\newblock {\em In Proceedings of the VLDB} 4(6).

\bibitem[\protect\citeauthoryear{Poon and Domingos}{2006}]{poon:aaai06}
Poon, H., and Domingos, P.
\newblock 2006.
\newblock Sound and efficient inference with probabilistic and deterministic
  dependencies.
\newblock In {\em National Conference on Artificial Intelligence}.

\bibitem[\protect\citeauthoryear{Qu, Bengio, and Tang}{2019}]{qu:icml19}
Qu, M.; Bengio, Y.; and Tang, J.
\newblock 2019.
\newblock Gmnn: Graph markov neural networks.
\newblock In {\em International Conference on Machine Learning}.

\bibitem[\protect\citeauthoryear{Qu \bgroup et al\mbox.\egroup
  }{2014}]{qu:ecml14}
Qu, Q.; Liu, S.; Jensen, C.~S.; Zhu, F.; and Faloutsos, C.
\newblock 2014.
\newblock Interestingness-driven diffusion process summarization in dynamic
  networks.
\newblock In {\em Joint European Conference on Machine Learning and Knowledge
  Discovery in Databases}.

\bibitem[\protect\citeauthoryear{Rajaraman and
  Ullman}{2011}]{rajaraman2011mining}
Rajaraman, A., and Ullman, J.~D.
\newblock 2011.
\newblock {\em Mining of massive datasets}.

\bibitem[\protect\citeauthoryear{Richardson and
  Domingos}{2006}]{richardson:ml06}
Richardson, M., and Domingos, P.
\newblock 2006.
\newblock Markov logic networks.
\newblock {\em Machine learning} 62:107--136.

\bibitem[\protect\citeauthoryear{Scott}{1988}]{john:soc88}
Scott, J.
\newblock 1988.
\newblock Social network analysis.
\newblock {\em Sociology} 22:109--127.

\bibitem[\protect\citeauthoryear{Sen \bgroup et al\mbox.\egroup
  }{2008}]{sen:ai08}
Sen, P.; Namata, G.; Bilgic, M.; Getoor, L.; Gallagher, B.; and Eliassi{-}Rad,
  T.
\newblock 2008.
\newblock Collective classification in network data.
\newblock {\em {AI} Magazine} 29:93--106.

\bibitem[\protect\citeauthoryear{Shi \bgroup et al\mbox.\egroup
  }{2015}]{shi:kde15}
Shi, L.; Tong, H.; Tang, J.; and Lin, C.
\newblock 2015.
\newblock Vegas: Visual influence graph summarization on citation networks.
\newblock {\em Knowledge and Data Engineering} 27:3417--3431.

\bibitem[\protect\citeauthoryear{Tan, Steinbach, and Kumar}{2006}]{tan:book06}
Tan, P.-N.; Steinbach, M.; and Kumar, V.
\newblock 2006.
\newblock {\em Introduction to Data Mining}.
\newblock Pearson Education.

\bibitem[\protect\citeauthoryear{Veli{\v{c}}kovi{\'{c}} \bgroup et
  al\mbox.\egroup }{2018}]{velickovic:iclr18}
Veli{\v{c}}kovi{\'{c}}, P.; Cucurull, G.; Casanova, A.; Romero, A.; Li{\`{o}},
  P.; and Bengio, Y.
\newblock 2018.
\newblock Graph attention networks.
\newblock {\em International Conference on Learning Representations}.

\bibitem[\protect\citeauthoryear{Wasserman and Faust}{1994}]{faust:book94}
Wasserman, S., and Faust, K.
\newblock 1994.
\newblock {\em Social Network Analysis: Methods and Applications}.
\newblock Cambridge University Press.

\bibitem[\protect\citeauthoryear{Wu \bgroup et al\mbox.\egroup
  }{2014}]{wu:iciseee14}
Wu, Y.; Zhong, Z.; Xiong, W.; and Jing, N.
\newblock 2014.
\newblock Graph summarization for attributed graphs.
\newblock In {\em International Conference on Information Science, Electronics
  and Electrical Engineering}.

\bibitem[\protect\citeauthoryear{Yang, Cohen, and
  Salakhutdinov}{2016}]{yang:icml16}
Yang, Z.; Cohen, W.~W.; and Salakhutdinov, R.
\newblock 2016.
\newblock Revisiting semi-supervised learning with graph embeddings.
\newblock In {\em International Conference on Machine Learning}.

\end{thebibliography}
\bibliographystyle{aaai}

\end{document}